\documentclass[letterpaper]{article} 
\usepackage{aaai2026}  
\usepackage{times}  
\usepackage{helvet}  
\usepackage{courier}  
\usepackage[hyphens]{url}  
\usepackage{graphicx} 
\usepackage{natbib}  
\usepackage{caption} 
\usepackage{algorithm}
\usepackage{algorithmic}
\usepackage{xcolor}
\usepackage{amsmath}
\usepackage{amssymb}
\usepackage{booktabs}
\usepackage{graphicx}
\usepackage{subcaption}

\usepackage{newfloat}
\usepackage{listings}
\DeclareCaptionStyle{ruled}{labelfont=normalfont,labelsep=colon,strut=off} 
\floatstyle{ruled}
\newfloat{listing}{tb}{lst}{}
\floatname{listing}{Listing}
\title{MIRNet: Integrating Constrained Graph-Based Reasoning with Pre-training for Diagnostic Medical Imaging}
\author {
    Shufeng Kong\textsuperscript{\rm 1,4},
    Zijie Wang\textsuperscript{\rm 1},
    Nuan Cui\textsuperscript{\rm 2},
    Hao Tang\textsuperscript{\rm 2},
    Yihan Meng\textsuperscript{\rm 2},
    Yuanyuan Wei\textsuperscript{\rm 1},
    Feifan Chen\textsuperscript{\rm 1},
    Yingheng Wang\textsuperscript{\rm 4},
    Zhuo Cai\textsuperscript{\rm 6},
    Yaonan Wang\textsuperscript{\rm 6},
    Yulong Zhang\textsuperscript{\rm 5},
    Yuzheng Li\textsuperscript{\rm 1},
    Zibin Zheng\textsuperscript{\rm 1},\\
    Caihua Liu\textsuperscript{\rm 3,4}\thanks{Corresponding authors: Caihua Liu (cl2869@cornell.edu), Hao Liang (lianghao@hnucm.edu.cn)},
    Hao Liang\textsuperscript{\rm 2*}
}
\affiliations {
    \textsuperscript{\rm 1}School of Software Engineering, Sun Yat-sen University, Zhuhai, China\\
    \textsuperscript{\rm 2}Institute of TCM Diagnostics, Hunan University of Chinese Medicine, Changsha, China\\
    \textsuperscript{\rm 3} School of Artificial Intelligence, Guilin University of Electronic Technology, Guilin, China \\
    \textsuperscript{\rm 4} Department of Computer Science, Cornell University, Ithaca, NY, USA \\
    \textsuperscript{\rm 5}The Fifth Affiliated Hospital, Sun Yat-sen University, Zhuhai, China \\
    \textsuperscript{\rm 6} Merchants Union Consumer Finance Company Limited (MUCFC), Shenzhen, China \\

}

\usepackage{bibentry}

\begin{document}

\maketitle

\begin{abstract}
Automated interpretation of medical images demands robust modeling of complex visual-semantic relationships while addressing annotation scarcity, label imbalance, and clinical plausibility constraints. We introduce MIRNet (Medical Image Reasoner Network), a novel framework that integrates self-supervised pre-training with constrained graph-based reasoning. Tongue image diagnosis is a particularly challenging domain that requires fine-grained visual and semantic understanding. Our approach leverages self-supervised masked autoencoder (MAE) to learn transferable visual representations from unlabeled data; employs graph attention networks (GAT) to model label correlations through expert-defined structured graphs; enforces clinical priors via constraint-aware optimization using KL divergence and regularization losses; and mitigates imbalance using asymmetric loss (ASL) and boosting ensembles. To address annotation scarcity, we also introduce TongueAtlas-4K, a comprehensive expert-curated benchmark comprising 4,000 images annotated with 22 diagnostic labels--representing the largest public dataset in tongue analysis. Validation shows our method achieves state-of-the-art performance. While optimized for tongue diagnosis, the framework readily generalizes to broader diagnostic medical imaging tasks. 

\end{abstract}

 \begin{links}
 \link{Code}{https://github.com/zijie8247/MIRNet}
\link{Datasets}{https://doi.org/10.5281/zenodo.17557646}
 \end{links}


\section{Introduction}

Medical image diagnosis requires recognizing intricate visual patterns with domain knowledge, and this process demands nuanced reasoning about statistically correlated diagnostic labels and clinical priors. In tongue analysis, for instance, ``pale tongue" frequently co-occurs with ``white tongue coating," yet such domain knowledge still remain underexplored in current approaches. Recent years have witnessed significant advances in automated tongue image diagnosis, yet critical limitations persist. \citet{jiang2022deep} proposed separate Residual Networks (ResNets) for individual tongue labels with late-stage output fusion, neglecting inter-label dependencies. While their work utilized a substantial dataset of 8,676 expert-annotated images covering seven categories (fissured, tooth-marked, stasis, spotted, greasy coating, peeled coating, rotten coating), this represents only a partial diagnostic spectrum, and the dataset remains non-public.  \citet{liu2024research} introduced LGAN, combining streamlined Convolutional Neural Networks (CNNs) with dual attention mechanisms (channel-wise + spatial) for feature extraction, followed by bidirectional Recurrent Neural Networks (RNNs) to model label correlations. Most recently, \citet{liang2025tongue} developed IF-RCNet, a two-tier architecture employing dilated convolutions for expanded receptive fields and residual convolutional block attention modules for feature fusion, enabling segmentation-classification synergy. Despite these innovations, no existing framework systematically addresses the following interconnected challenges: (1) annotation scarcity hindering supervised learning in specialized domains, (2) severe label imbalance skewing performance toward prevalent conditions, (3) inadequate label correlation modeling limiting diagnostic coherence, and (4) unconstrained predictions generating clinically implausible outcomes.

In scientific domains including biomedicine, integrating learning with reasoning has become increasingly vital --- enhancing data efficiency, improving pattern recognition, and yielding scientifically valid outcomes. Exemplifying this synergy: Deep Reasoning Networks (DRNets) merge deep learning with constraint optimization to embed thermodynamic priors for automated material discovery \cite{chen2021automating}, enabling accurate phase mapping of crystal mixtures with limited unlabeled data; Physics-Informed Neural Networks (PINNs) encode governing differential equations directly into neural architectures \cite{cuomo2022scientific}, penalizing PDE violations during training to achieve robust solutions in data-scarce engineering and biophysical applications. Yet in diagnostic medical imaging, particularly tongue analysis, this paradigm of integrating domain knowledge with data-driven learning remains critically under explored, leaving substantial potential for improved diagnostic coherence untapped.

To bridge this gap, we propose MIRNet, an end-to-end architecture to integrate constrained graph-based reasoning with pre-training for tongue image diagnosis. First, our MAE Visual Encoder \cite{he2022masked} adapts Vision Transformers to medical domains through self-supervised reconstruction of masked anatomical regions, learning transferable representations from unlabeled data. Second, the Constrained GAT Decoder \cite{velickovic2018graph} processes expert-defined label graphs—where nodes represent diagnostic labels and edges encode statistical co-occurrences, while enforcing clinical plausibility through a custom loss term encodes domain knowledge such as physiological incompatibilities (e.g., ``thin tongue" excludes ``enlarged tongue"). Third, joint optimization with asymmetric loss handles label imbalance by down-weighting negative gradients for prevalent classes. Fourth, boosting ensembles iteratively refine predictions to enhance robustness. We evaluate MIRNet on tongue image diagnosis using TongueAtlas-4K, a comprehensive benchmark curated by medical experts. This dataset contains 4,000 images annotated with 22 clinically validated labels spanning tongue color, tongue shape,  property of tongue coating, and color of tongue coating.

Our work delivers four key contributions to tongue image diagnosis and medical AI:
\begin{itemize}
    \item \textbf{MIRNet}: A pioneering framework that \textbf{integrates self-supervised visual pre-training with constrained graph reasoning}, addressing annotation scarcity while modeling diagnostic dependencies through clinical knowledge graphs. 
    
    \item \textbf{TongueAtlas-4K}: The largest \textbf{publicly available expert-curated benchmark} for tongue analysis, featuring 4,000 images annotated with 22 clinically validated labels spanning color, texture, and morphology to accelerate community research.
    
    \item \textbf{Differentiable clinical constraint engine}: A novel constraint-aware optimization engine using \textbf{KL-divergence and domain-driven regularization losses} to encode medical knowledge (e.g., physiological incompatibilities) as soft constraints and thus reduce implausible predictions. The overall system further employs asymmetric loss (ASL) to mitigate label imbalance.
    
    \item \textbf{State-of-the-art performance}: 
    Our model consistently outperforms all baselines across multiple metrics, improving Macro Recall by \textbf{77.8\%} and Macro-F1 by \textbf{33.2\%} over the strongest competing method. Ablation studies further validate the effectiveness of our proposed components.
\end{itemize}

\section{Preliminaries}
This section formalizes the setting of multi‑label medical image diagnosis and highlights its characteristics.




\subsection{Problem Formulation}
Let $\mathcal{X} = \mathbb{R}^{H \times W \times C}$ denote the medical image space and $\mathcal{Y} = \{0,1\}^K$ the label space for $K$ distinct diagnoses. Given an image $\mathbf{X} \in \mathcal{X}$, we aim to learn a mapping $f_\theta: \mathcal{X} \to \mathcal{Y}$ that predicts a label vector $\mathbf{y} = (y_1, \dots, y_K)^\top$ satisfying clinical knowledge constraints while accounting for data distribution characteristics:

\begin{enumerate}
    \item \textbf{Clinical Constraints}: The prediction must satisfy a set of domain knowledge rules $\Phi = \{\phi_j\}_{j=1}^m$ where $\phi_j: \mathcal{Y} \to \{0,1\}$ is defined as:
    \begin{align*}
        \phi_j(\mathbf{y}) = 0 \quad &\iff \quad \text{constraint } j \text{ is satisfied} \\
        \phi_j(\mathbf{y}) = 1 \quad &\iff \quad \text{constraint violation}
    \end{align*}
    with representative constraints:
    \begin{itemize}
        \item \textit{Mutual exclusion}: $\phi_j(\mathbf{y}) = y_a \cdot y_b$ \hfill (e.g., diagnoses $a$ and $b$ cannot co-occur)
        \item \textit{Co-appearance}: $\phi_j(\mathbf{y}) = |y_a - y_b|$ \hfill (e.g., $a$ and $b$ must both be present or both absent)
        \item \textit{Implication}: $\phi_j(\mathbf{y}) = y_a \cdot (1 - y_b)$ \hfill (e.g., $a$ requires presence of $b$)
    \end{itemize}

    \item \textbf{Label Imbalance}: The data exhibits significant class imbalance where $\exists k \in \{1,\dots,K\}$ such that $\mathbb{P}(y_k=1) \leq \tau$ with $\tau \ll 0.5$.

    \item \textbf{Label Dependencies}: Diagnoses exhibit statistical dependencies characterized by non-zero off-diagonal covariance $\Sigma_{ij} = \text{Cov}(y_i, y_j) \neq 0$ for some $i \neq j$.
\end{enumerate}

A model for the problem must simultaneously: 
\begin{itemize}
    \item Guarantee clinical plausibility: $\phi_j(f_\theta(\mathbf{X})) = 0 \quad \forall j$
    \item Maintain robustness under class imbalance ($\tau \ll 0.5$)
    \item Exploit statistical dependencies ($\Sigma \neq \mathbf{I}_K$)
\end{itemize}

Therefore, for effective tongue image diagnosis, we seek to develop a model $f_\theta$ that simultaneously:
(i) Addresses the data annotation scarcity,
(ii) Enforces strict adherence to clinical constraints $\Phi$ through constraint-aware optimization,
(iii) Maintains robustness under severe class imbalance ($\tau \ll 0.5$), and
(iv) Exploits statistical label dependencies ($\Sigma \neq \mathbf{I}K$).
This requires integrated utilization of clinical priors (via $\Phi$) and statistical priors (label distributions) during training, while ensuring all predictions satisfy $\phi_j(f\theta(\mathbf{X})) = 0$ $\forall j \in \{1,\dots,m\}$.

\begin{figure*}[htbp]
    \centering
    \includegraphics[width=0.90\linewidth]{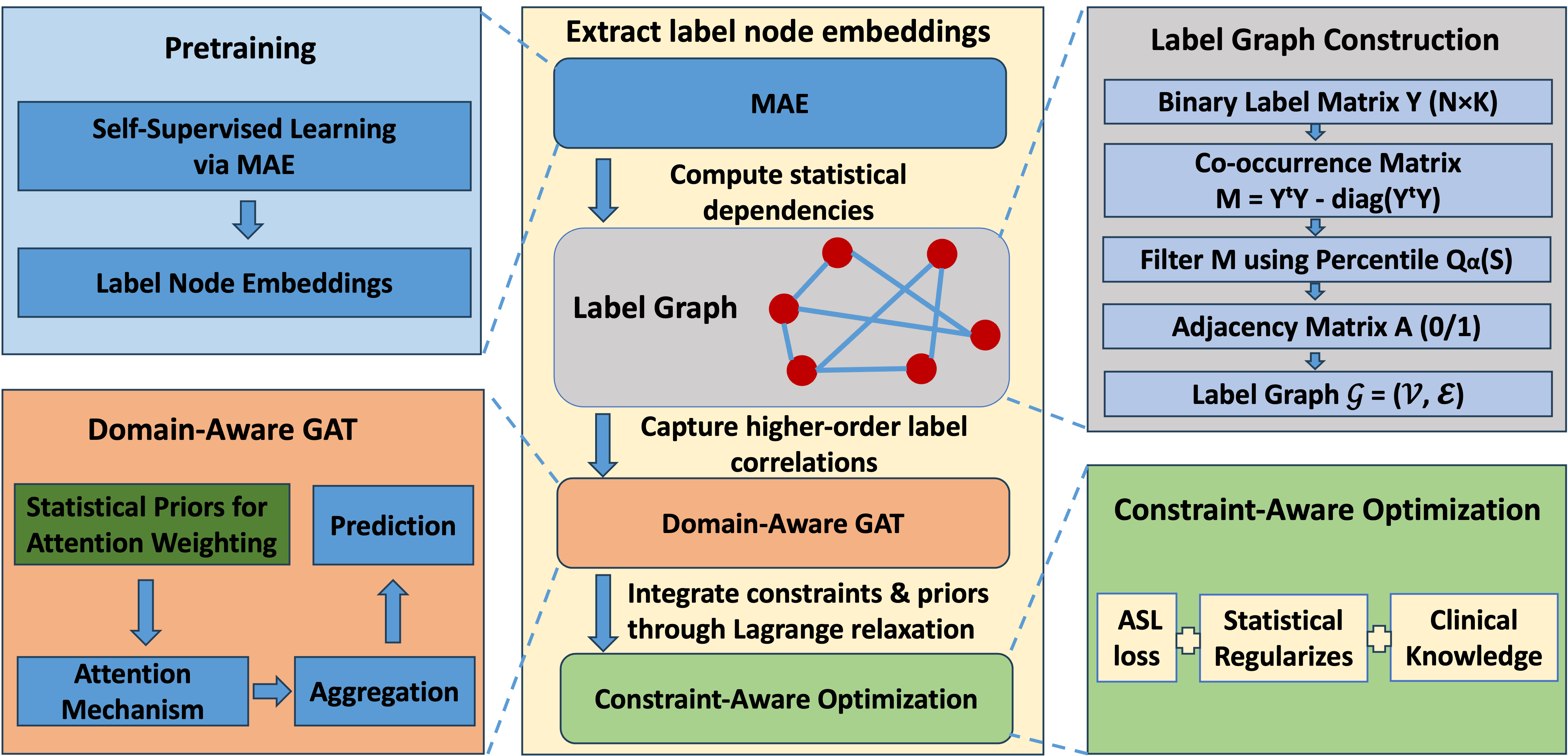}
    \caption{The overall architecture of MIRNet. The central diagram shows the main workflow: a pretrained MAE extracts image embeddings, a label graph is built from statistical dependencies, and a domain-aware GAT captures higher-order label correlations. The model is then trained via a constraint-aware optimization mechanism. The left panel details domain-aware pretraining and the GAT, while the right panel illustrates label graph construction and constraint-aware optimization.}
    \label{fig:arch}
\end{figure*}

\section{Methodology}\label{sec:methodology}
In this section, we introduce MIRNet. The overall architecture is shown in Figure~\ref{fig:arch}, and each component is detailed in the following subsections.
\subsection{Masked Autoencoder Pretraining}

To address the limited availability of annotated medical images, we adopt Masked Autoencoder (MAE) pretraining \cite{he2022masked} on large-scale unlabeled tongue image data. MAE is a self-supervised learning paradigm that reconstructs missing visual content from partial observations, enabling robust feature extraction suitable for downstream diagnostic tasks.

\subsubsection{Pretraining Workflow}
Given an unlabeled input image \(\mathbf{X} \in \mathbb{R}^{H \times W \times C}\), the MAE workflow proceeds as follows:

\begin{enumerate}
    \item \textbf{Patch Partitioning.}  
    Divide \(\mathbf{X}\) into \(N\) non-overlapping patches:  
    \[
    \mathbf{X} = [\mathbf{x}_1, \mathbf{x}_2, \dots, \mathbf{x}_N], \quad \mathbf{x}_i \in \mathbb{R}^{P \times P \times C}.
    \]

    \item \textbf{Random Masking.}  
    Generate a binary mask \(\mathbf{M} \in \{0,1\}^N\) with a high masking ratio \(\rho = 0.75\), yielding visible patch set:
    \[
    \mathbf{X}_{\text{vis}} = \{\mathbf{x}_i \mid M_i = 1\}, \quad N_v = (1 - \rho)N.
    \]

    \item \textbf{Encoder Processing.}  
    A Vision Transformer (ViT) encoder \(f_{\text{enc}}\) maps the embedded visible patches:
    \[
    \mathbf{H} = f_{\text{enc}}(\mathbf{E}\mathbf{X}_{\text{vis}} + \mathbf{P}),
    \]
    where \(\mathbf{E}\) is the patch embedding matrix and \(\mathbf{P}\) positional encodings.

    \item \textbf{Decoder Reconstruction.}  
    A lightweight transformer decoder \(f_{\text{dec}}\) reconstructs masked patches using encoder outputs and mask tokens:
    \[
    \hat{\mathbf{X}} = f_{\text{dec}}([\mathbf{H}, \mathbf{T}_{\text{mask}}]).
    \]

    \item \textbf{Reconstruction Loss.}  
    The model minimizes the pixel-wise mean squared error (MSE) over masked regions:
    \[
    \mathcal{L}_{\text{MAE}} = \frac{1}{|\mathcal{M}|} \sum_{i \in \mathcal{M}} \|\mathbf{x}_i - \hat{\mathbf{x}}_i\|_2^2,
    \quad \mathcal{M} = \{i \mid M_i = 0\}.
    \]
\end{enumerate}

\subsubsection{Transfer Learning Protocol}
The MAE-pretrained encoder serves as the foundational visual front-end for our diagnostic framework. Its general image features initialize node embeddings in the subsequent Graph-Based Label Correlation Modeling, establishing a visual-semantic prior that bridges low-level image patterns with high-level diagnostic concepts. This creates a unified pipeline where: (1) Anatomically relevant regions within the extracted visual features ground diagnostic predictions, and (2) Graph propagation refines these predictions by leveraging statistical label dependencies. The pretrained encoder thus provides the core visual representation for the comprehensive diagnostic system detailed in later sections.

\subsection{Graph-Based Label Correlation Modeling}
We extend the standard GAT to a framework tailored for diagnostic label modeling. Our approach operates on a label graph \(\mathcal{G} = (\mathcal{V}, \mathcal{E})\), where:
\begin{itemize}
    \item Nodes \(\mathcal{V} = \{v_1, \dots, v_K\}\) represent diagnostic labels.
    \item Edges \(\mathcal{E}\) encode statistically significant co-occurrence.
\end{itemize}

\subsubsection{Label Graph Construction}
Define the binary label matrix $\mathbf{Y} \in \{0,1\}^{N \times K}$ for $N$ samples and $K$ diagnoses. The empirical co-occurrence matrix is:
\[
\mathbf{M} = \mathbf{Y}^\top \mathbf{Y} - \text{diag}(\mathbf{Y}^\top \mathbf{Y})
\]
where off-diagonal elements $M_{ij}$ count co-occurrences between labels $i$ and $j$. The adjacency matrix $\mathbf{A}$ is obtained via dynamic thresholding:
\[
\mathbf{A}_{ij} = 
\begin{cases} 
1 & \text{if } M_{ij} \geq Q_{\alpha}(\mathcal{S}) \\
0 & \text{otherwise}
\end{cases}, \quad \mathcal{S} = \{M_{ij} > 0\}
\]
with $Q_{\alpha}$ being the $\alpha$-th percentile ($\alpha=25$) of non-zero co-occurrences. This yields a sparse graph $\mathcal{G} = (\mathcal{V}, \mathcal{E})$ where nodes $\mathcal{V} = \{1,\dots,K\}$ represent labels.

\subsubsection{Label Propagation}

We propagate label embeddings through \(L\) GAT layers, thereby capturing higher‑order (multi‑hop) label correlations:

\begin{enumerate}
  \item \textbf{Initialization.}  
    Set node features to visual embeddings from the MAE encoder:  
    \[
      \mathbf{v}_i^{(0)} = \mathbf{z}_i \in \mathbb{R}^d.
    \]

  \item \textbf{Attention Mechanism.}  
    At layer \(l\), compute attention coefficients for each neighbor \(j \in \mathcal{N}(i)\):
    \[
      e_{ij}^{(l)}
      = \mathrm{LeakyReLU}\bigl(\mathbf{a}^{(l)\top}
        [\,W^{(l)}\mathbf{v}_i^{(l)} \,\Vert\, W^{(l)}\mathbf{v}_j^{(l)}]\bigr),
    \]
    \[
      \alpha_{ij}^{(l)}
      = \frac{\exp\bigl(e_{ij}^{(l)}\bigr)}
             {\sum_{k \in \mathcal{N}(i)} \exp\bigl(e_{ik}^{(l)}\bigr)}.
    \]

  \item \textbf{Aggregation.}  
    Update each node by attending to its neighbors:
    \[
      \mathbf{v}_i^{(l+1)}
      = \sigma\!\Bigl(\sum_{j \in \mathcal{N}(i)}
          \alpha_{ij}^{(l)}\,W^{(l)}\,\mathbf{v}_j^{(l)}\Bigr),
    \]
    where \(\sigma(\cdot)\) is a nonlinearity (e.g., ReLU).

\item \textbf{Prediction Head.}
  To produce final diagnostic probabilities, we fuse the original visual embedding with the graph‑refined representation for each label \(k\):
\[
\hat{y}_k = \sigma\!\Bigl(\mathbf{w}_k^\top \bigl[\mathbf{v}_k^{(0)} \,\Vert\, \mathbf{v}_k^{(L)}\bigr] + b_k\Bigr),
\]
where
\begin{itemize}
  \item \(\mathbf{v}_k^{(0)}\) is the initial MAE-derived visual feature for label \(k\),
  \item \(\mathbf{v}_k^{(L)}\) is the corresponding output after \(L\) GAT layers,
  \item \(\Vert\) denotes vector concatenation,
  \item \(\mathbf{w}_k\in\mathbb{R}^{2d'}\) and \(b_k\) are learnable classification parameters,
  \item \(\sigma(\cdot)\) is the sigmoid activation, yielding \(\hat{y}_k\in(0,1)\).
\end{itemize}

This design preserves localized visual evidence while enriching it with context‑aware label correlations.  
\end{enumerate}

\subsubsection{Diagnostic-Specific Enhancements}

To better address clinically rare conditions and emphasize strong empirical co-occurrences, we augment each GAT layer with two mechanisms:

\begin{itemize}
  \item \textbf{Rare Label Boosting.}  
    Increase the influence of under-represented labels \(k\) by re-scaling their outgoing attention:
    \[
      \alpha_{kj}
      \;\longleftarrow\;
      \alpha_{kj}\,\Bigl(1 + \log\tfrac{1}{\mathbb{P}(y_k=1)}\Bigr).
    \]

  \item \textbf{Correlation Confidence Weighting.}  
    Weight each edge’s attention by its normalized co-occurrence frequency:
    \[
      \tilde{\alpha}_{ij}
      = \alpha_{ij}\;\frac{M_{ij}}{\max_{u,v} M_{uv}}.
    \]
\end{itemize}

By integrating these enhancements directly into the attention computation, our model adaptively balances rare-label emphasis against empirical co-occurrence strength, yielding a transparent, data-driven mechanism for capturing clinically relevant label interdependencies.

\subsection{Constraint-Aware Optimization}
We develop a unified optimization framework integrating clinical constraints and statistical priors through Lagrange relaxation:

\begin{align}
\min_{\theta, \phi}\; 
&\underbrace{\mathcal{L}_{\text{ASL}}\bigl(f_\theta(g_\phi(\mathbf{X})),\,\mathbf{y}\bigr)}_{\substack{\text{diagnosis loss}}} 
+ \lambda_1
\underbrace{\mathcal{L}_{\text{constraint}}}_{\substack{\text{clinical}\\[-0.2ex]\text{knowledge}}} 
+ \lambda_2
\underbrace{\mathcal{L}_{\text{prior}}}_{\substack{\text{statistical}\\[-0.2ex]\text{regularizers}}}
\label{eq:total_loss}
\end{align}

\begin{enumerate}

\item\textbf{Diagnostic Loss:}
Addresses label imbalance (\(\mathbb{P}(y_k=1) \leq \tau \ll 0.5\)) via Asymmetric Loss:
\[
\begin{aligned}
\mathcal{L}_{\text{ASL}} ={}& -\sum_{k=1}^{K} \gamma_{k} \Bigl[
  y_{k}\,(1 - p_{k})^{\zeta_{+}}\,\log p_{k} \\
  &\quad +\; (1 - y_{k})\,p_{k}^{\zeta_{-}}\,\log(1 - p_{k})
\Bigr]
\end{aligned}
\]
\begin{itemize}
  \item \(p_k = \sigma(z_k)\): predicted probability for class \(k\)
  \item \(\gamma_k = \sqrt{\tau/\mathbb{P}(y_k=1)}\): frequency-based re-weighting
  \item \(\zeta_+ < \zeta_-\): asymmetric focusing parameters
\end{itemize}

\item\textbf{Clinical Knowledge Constraints:}
Clinical rules \(\Phi = \{\phi_j\}_{j=1}^m\) are incorporated via Lagrange relaxation:
\begin{align*}
\mathcal{L}_{\text{constraint}} = \sum_{j=1}^m \mathbb{E}_{\mathbf{X}} \left[ \max\left(0, \phi_j(f_\theta(\mathbf{X}))\right) \right]
\end{align*}
where constraint functions \(\phi_j\) implement clinical rules:
\begin{itemize}
  \item \textit{Mutual exclusion}: \(\phi_j(\mathbf{p}) = p_a \cdot p_b\) \hfill (diagnoses \(a\), \(b\) cannot co-occur)
  \item \textit{Co-appearance}: \(\phi_j(\mathbf{p}) = |p_a - p_b|\) \hfill (\(a\) and \(b\) must both present/absent)
  \item \textit{Implication}: \(\phi_j(\mathbf{p}) = p_a \cdot (1 - p_b)\) \hfill (\(a\) requires presence of \(b\))
\end{itemize}
The \(\max(0, \cdot)\) operator ensures penalty only on constraint violations.

\item\textbf{Statistical Priors:}
\begin{align*}
\mathcal{L}_{\text{prior}} &= \text{KL}\big(q(\mathbf{y}|\mathbf{X}) \parallel p_{\text{data}}(\mathbf{y})\big) \\
&= \sum_{k=1}^K \pi_k \log \frac{\pi_k}{q_k}, \quad 
q_k = \mathbb{E}_{\mathbf{X}}[p_k(\mathbf{X})]
\end{align*}
where \(\pi_k = \mathbb{P}(y_k=1)\) is the empirical class prior.
\end{enumerate}

Overall, the framework attempts to address several challenges: 
(1) The $\max(0, \phi_j)$ formulation offer a differentiable approximation of clinical constraints, enabling gradient-based optimization while encouraging clinically plausible outputs; 
(2) ASL's combination of class re-weighting ($\gamma_k$) and asymmetric focusing ($\zeta_+, \zeta_-$) help mitigate extreme label imbalance by emphasizing rare positive cases; 
(3) KL-divergence regularization assist in aligning predictions with empirical class priors ($\pi_k$), possibly reducing marginal probability shift; and 
(4) The hyperparameters $\lambda_1$ and $\lambda_2$ offer a mechanism to balance clinical constraint satisfaction against statistical prior alignment within the unified objective.

\section{Experiments}

\subsection{Datasets}

We curated a large-scale open source tongue image dataset under the guidance of clinical experts, encompassing 4 diagnostic dimensions: tongue color, tongue shape,  property of tongue coating (physical characteristics), and color of tongue coating. Table~\ref{tab:tongue_terms} summarizes the 22 fine-grained classes defined across these dimensions \cite{ISO23961_1_2021} and their distribution across annotated images.


\begin{table}[ht]
\centering
\renewcommand{\arraystretch}{1.05} 
\begin{tabular}{clc}
\toprule
\textbf{Label} & \textbf{Tongue diagnosis term} & \textbf{Percentage} \\
\midrule
\multicolumn{3}{l}{\emph{Tongue color}} \\
0 & Pale tongue            & 23.67\% \\
1 & Light-red tongue       & 52.80\% \\
2 & Red tongue             & 14.27\% \\
3 & Dark-red tongue        & 2.15\%  \\
4 & Blue-purple tongue     & 15.53\% \\
\midrule
\multicolumn{3}{l}{\emph{Tongue shape}} \\
5  & Tender tongue                 & 4.60\%  \\
6  & Tough tongue                  & 4.70\%  \\
7  & Thin tongue                   & 8.05\%  \\
8  & Enlarged tongue               & 13.15\% \\
9  & Tongue with spots or thorns   & 22.88\% \\
10 & Tongue with cracks            & 21.57\% \\
11 & Tongue with teeth marks       & 53.67\% \\
\midrule
\multicolumn{3}{l}{\emph{Property of tongue coating}} \\
12 & No tongue coating             & 3.70\%  \\
13 & Peeled tongue coating         & 3.52\%  \\
14 & Thin tongue coating           & 67.58\% \\
15 & Thick tongue coating          & 24.12\% \\
16 & Moist tongue coating          & 46.08\% \\
17 & Dry tongue coating            & 5.55\%  \\
18 & Rotten and greasy tongue coating & 28.25\% \\
\midrule
\multicolumn{3}{l}{\emph{Color of tongue coating}} \\
19 & White tongue coating          & 78.38\% \\
20 & Yellow tongue coating         & 32.55\% \\
21 & Gray-black tongue coating     & 3.35\%  \\
\bottomrule
\end{tabular}
\caption{Label dimensions for tongue diagnosis terms and distribution across 4,000 images.}
\label{tab:tongue_terms}
\end{table}

\begin{table*}[ht!]
\centering
\begin{tabular}{lcccccc}
\toprule
Models & Example-F1 & Micro-F1 & Macro-F1 & Macro Precision & Macro Recall & Macro PR-AUC \\
\midrule
LGAN & 0.633779 & 0.640308 & 0.397091 & 0.504646 & 0.368678 & 0.492223 \\
YOLO12-CLS & 0.583403 & 0.591335 & 0.290485 & 0.379461 & 0.275153 & 0.400615 \\
Faster R-CNN & 0.650968 & 0.661723 & 0.380543 & 0.485094 & 0.338691 & 0.493433 \\
IFRCNet & 0.563874 & 0.567665 & 0.245823 & 0.314741 & 0.245877 & 0.491997 \\
DenseNet121 & 0.648428 & 0.657014 & 0.403075 & 0.487452 & 0.363772 & 0.350961 \\
C-GMVAE & 0.634429 & 0.646858 & 0.346378 & 0.459010 & 0.304918 & 0.526044 \\
\midrule
MIRNet & \textbf{0.680389} & 0.\textbf{683048} & 0.525425 & \textbf{0.507837} & 0.599019 & 0.527103 \\
MIRNet-Boosting & 0.674805 & 0.677620 & \textbf{0.537061} & 0.499404 & \textbf{0.655388} & \textbf{0.543415} \\
\bottomrule
\end{tabular}
\caption{Performance comparison of all considered models. Bold values indicate the best results. Each experiment was run five times, and the mean performance is reported.}
\label{tab:models_performance}
\end{table*}

\begin{figure*}[htbp]
  \centering
  \begin{subfigure}[b]{0.49\textwidth}
    \includegraphics[width=\linewidth]{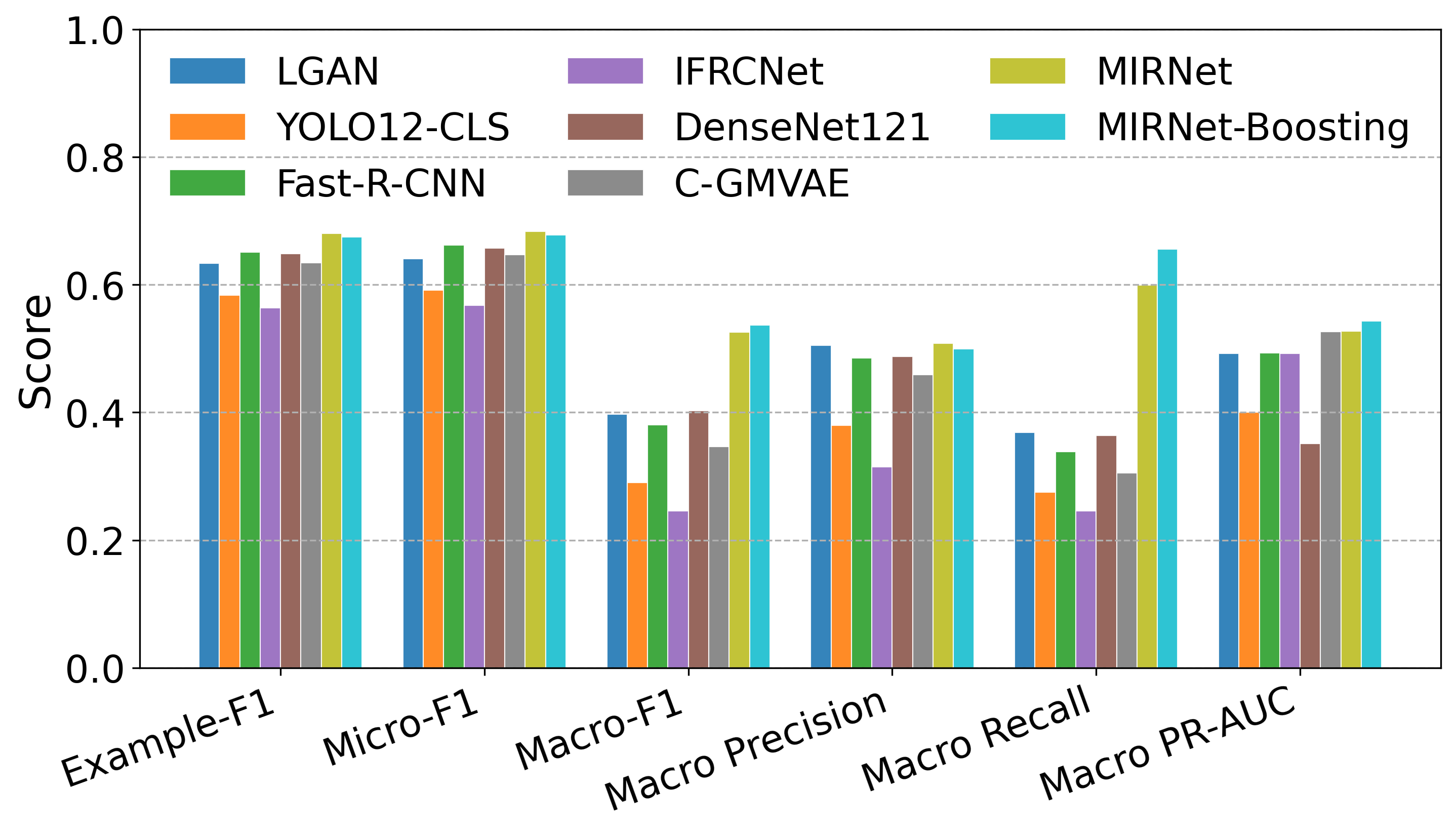}
  \end{subfigure}
  \hfill
  \begin{subfigure}[b]{0.50\textwidth}
    \includegraphics[width=\linewidth]{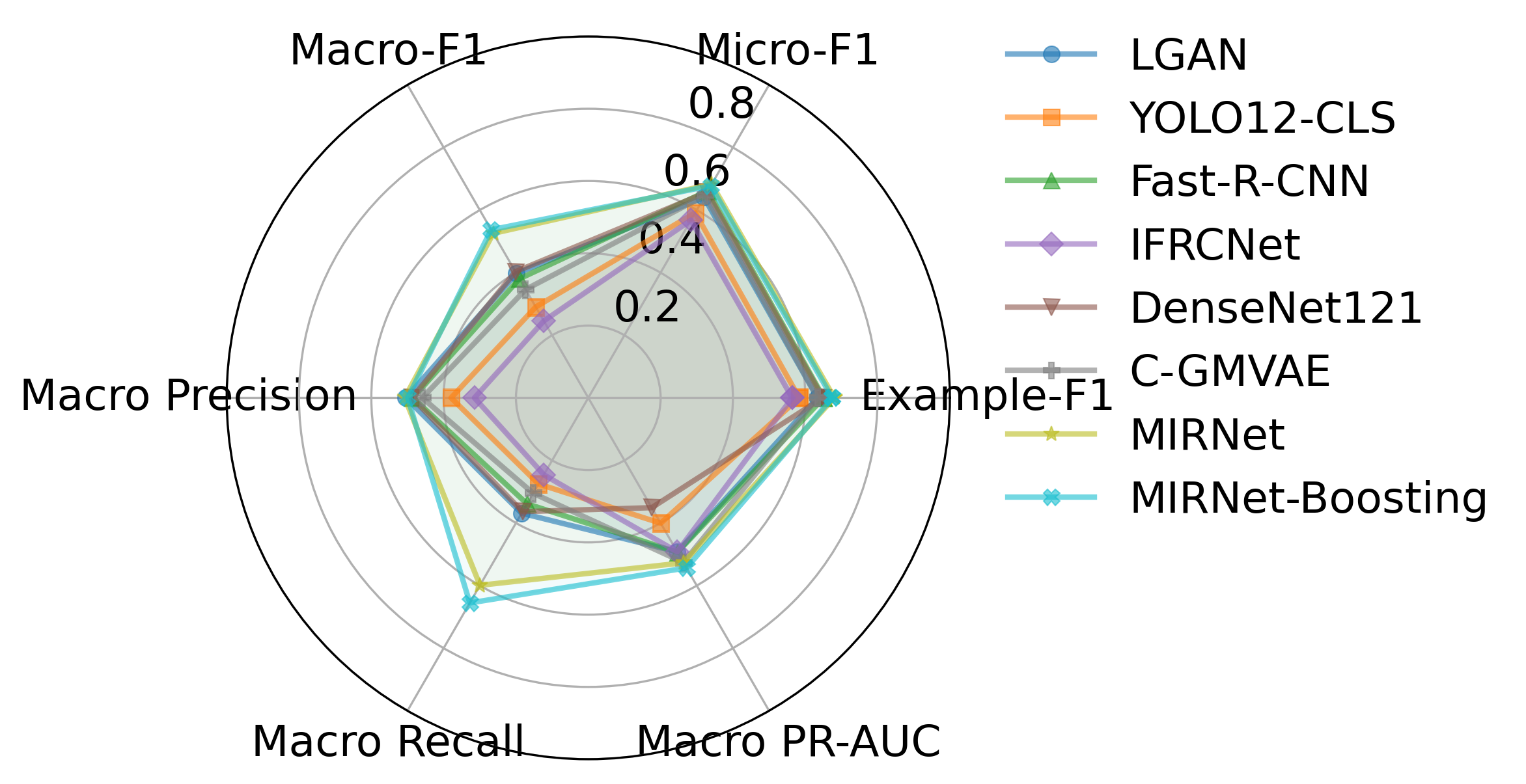}
  \end{subfigure}
  \caption{Visual comparison of all considered models using bar and radar charts.}
  \label{fig:model_comparison}
\end{figure*}

The annotated dataset comprises 4,000 tongue images collected from two independent clinical sources. Additionally, a total of 15,905 unlabeled images
were curated to support large-scale pretraining.

All annotated images underwent a consensus-based annotation pipeline: ten systematically trained experts independently labeled the samples, followed by mutually blinded cross-review. Discrepancies were resolved through dual expert audits and finalized through adjudication by a senior traditional medicine practitioner. The annotated dataset was then randomly split into 80\% training, 10\% validation, and 10\% test sets.

To enhance visual feature learning, all images were  preprocessed via rigorous color correction protocols—this included haze removal for affected images and reflectance normalization for clinically captured samples. Tongue regions were first segmented using DeepLabV3+ \cite{chen2018encoder}, and then followed by manual refinements using the ITK-Snap tool \cite{yushkevich2016itk}, ensuring precise anatomical representation.

\subsection{Experimental Settings}
\subsubsection{Baselines}
We benchmark our approach against three state-of-the-art tongue image diagnostic baselines, as introduced earlier: \textbf{Faster R-CNN} \cite{jiang2022deep}, \textbf{LGAN} \cite{liu2024research}, and \textbf{IFRCNet} \cite{liang2025tongue}.
To provide broader context, we also evaluate three general-purpose classification models on our dataset:
\begin{itemize}
    \item \textbf{DenseNet‑121}: A widely used convolutional network that has achieved state‑of‑the‑art results in past multi‑label medical image classification tasks (e.g. tongue‑based disease detection).
    \item \textbf{YOLO12‑CLS}: A classification branch adapted from YOLO detection architectures; its strong feature extraction capability make it a useful baseline for tongue image classification.
    \item \textbf{C-GMVAE}: A contrastive learning–boosted multi-label prediction model based on a Gaussian Mixture Variational Autoencoder. C‑GMVAE excels at modeling label correlations while learning latent space alignment for both features and labels.
\end{itemize}
All baseline implementations use original authors' codebases adapted to our dataset, and the hyperparameters for all models we evaluated, including our model MIRNet, were optimized using grid search. 

\subsubsection{Model Architecture}

Our proposed model integrates a Vision Transformer (ViT)-based encoder with a GAT decoder and a multi-layer perceptron (MLP) classifier head. 
The ViT backbone (\texttt{ViT-Base-Patch16-224}) was pretrained using the MAE framework on 19,505 unlabeled images. 
Pretraining employed a 75\% patch masking strategy with pixel-wise mean squared error (MSE) loss, configured with the following hyperparameters: \texttt{patch\_size=16}, \texttt{embed\_dim=768}, \texttt{depth=12}, \texttt{num\_heads=12}, \texttt{mlp\_ratio=4}, \texttt{qkv\_bias=True}, and \texttt{norm\_layer=partial (nn.LayerNorm, eps=1e-6)}. The overall pretraining workflow is detailed in the Methodology section.

\subsubsection{Fine-Tuning} 
During fine-tuning, the pretrained ViT encoder is augmented with a two-layer \texttt{GATv2Conv} module \cite{Fey/Lenssen/2019} to model inter-label dependencies. 
This module utilizes a domain-specific co-occurrence graph constructed from training set statistics and refined with clinical prior knowledge to strengthen correlated label pairs while suppressing mutually exclusive relationships. 
The \texttt{GATv2Conv} hyperparameters are: \texttt{in\_dim=768}, \texttt{hidden\_dim=64}, \texttt{out\_dim=21}, and \texttt{num\_head=8}. 
Final per-label predictions are generated by a shared two-layer MLP classifier with parameter dimensions $640 \times 320$ and $320 \times 1$, employing ReLU activation.

\subsubsection{Training Protocol} 
Optimization minimizes the composite loss defined in Equation~\eqref{eq:total_loss} with coefficients $\lambda_1 = 0.1$ and $\lambda_2 = 0.05$. 
Training employs the AdamW optimizer with a base learning rate of $1\times10^{-3}$, batch size of 200, and layer-wise decay ($\mathrm{layer\_decay} = 0.75$) over 200 epochs. 
All experiments were executed on an NVIDIA A800 GPU with cuDNN acceleration.

\subsubsection{MIRNet-Boosting} 
To further address minority class underperformance (F1-score $< 0.5$), we implement a dual-model boosting strategy:
\begin{itemize}
    \item A base model is trained on the full dataset;
    \item A second model is fine-tuned exclusively on underperforming classes using RandAugment, random erasing, and normalization for data augmentation.
\end{itemize}
Final predictions combine outputs from both models: The five lowest-performing labels are replaced by the second model's predictions, while all other labels retain the base model's outputs.

\begin{figure*}[htbp]
  \centering
  \begin{subfigure}[b]{0.74\textwidth}
    \includegraphics[width=\linewidth]{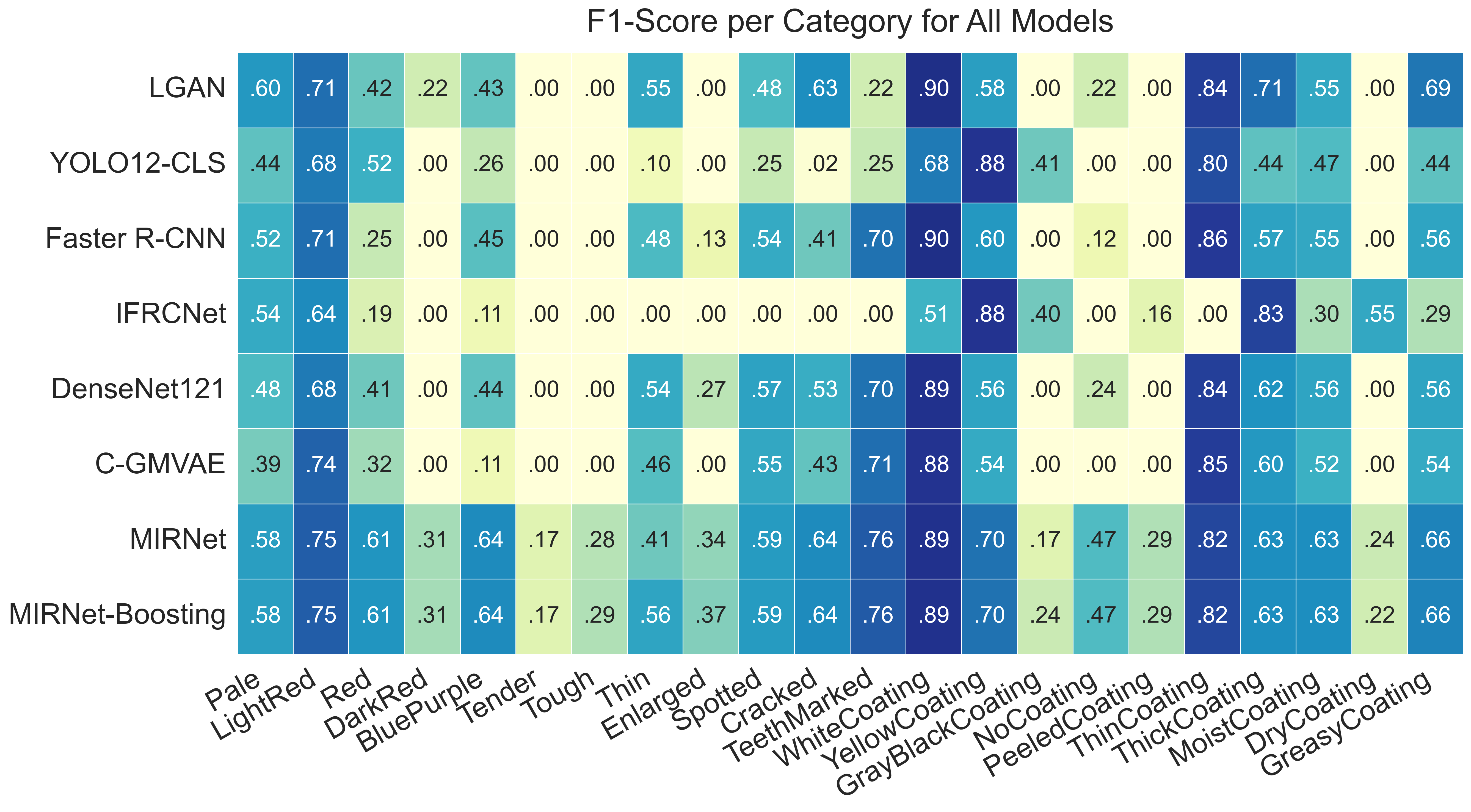}
  \end{subfigure}
  \hfill
  \begin{subfigure}[b]{0.245\textwidth}
    \includegraphics[width=\linewidth]{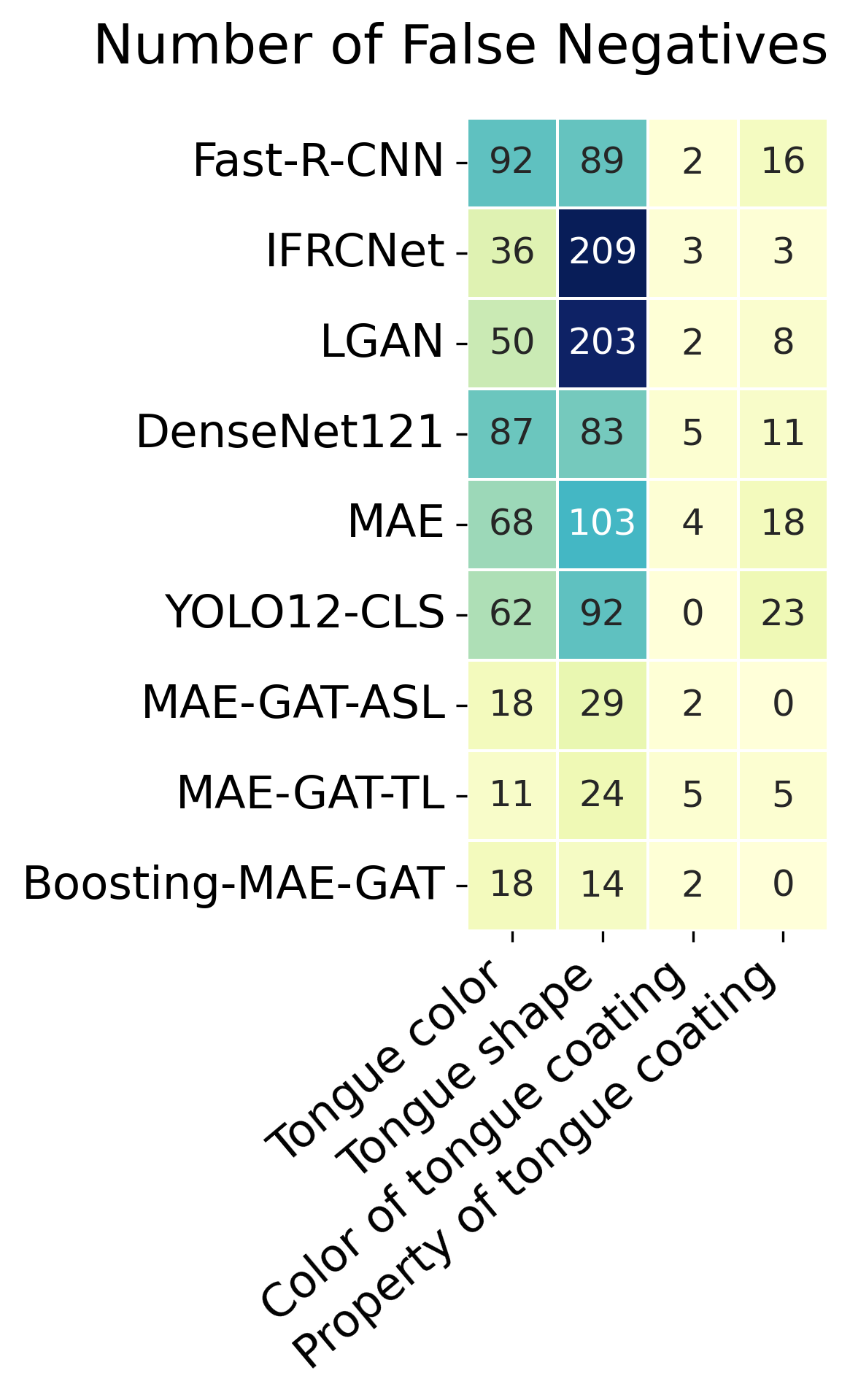}
  \end{subfigure}
  \caption{Diagnostic capability assessment: (Left) F1-score distribution heatmap across 22 tongue subcategories; (Right) Dimension-level missed-detection heatmap.}
  \label{fig:model_comparison1}
\end{figure*}

\subsection{Experimental Results and Analysis}


We evaluated all models using standard multi-label classification metrics: Example-F1, Micro-F1, Macro-F1, Macro Precision, Macro Recall, and Macro PR-AUC (Precision-Recall Area Under the Curve). Results in Table~\ref{tab:models_performance} demonstrate that \textbf{MIRNet} and \textbf{MIRNet-Boosting} consistently outperform all baselines across every metric. Crucially, MIRNet-Boosting achieves state-of-the-art performance in \textbf{Macro-F1 (0.537)}, \textbf{Macro Recall (0.655)}, and \textbf{Macro PR-AUC (0.543)}, improving Macro Recall by \textbf{77.8\%} and Macro-F1 by \textbf{33.2\%} over the strongest baseline). Even without boosting, MIRNet alone surpasses all baselines with \textbf{62.5\% higher Macro Recall} and \textbf{30.4\% higher Macro-F1}. Visual comparisons in Figure~\ref{fig:model_comparison} further highlight these performance gaps.

The exceptional Macro-Recall improvements (62.5--77.8\%) demonstrate MIRNet's sensitivity to rare classes, critical given the severe label imbalance shown in Table~1. These gains stem from three synergistic components: the Asymmetric Loss down-weights negative gradients for frequent classes to amplify rare positives; the boosting ensemble directly targets underperforming labels through dedicated fine-tuning; and the GAT's Rare Label Boosting rescales attention weights using inverse class frequency.

The substantial Macro-F1 gains (30.4--33.2\%) show that MIRNet achieves balanced precision and recall across all 22 labels, which is critical given the multifaceted classification challenge. These gains stem from two synergistic advantages: MAE pretraining leverages 15,905 unlabeled images to overcome annotation scarcity that handicaps tongue-specific baselines (e.g., LGAN); and integrated reasoning combines explicit label dependencies with clinical constraints that general models (e.g., C-GMVAE) inherently lack, proving domain adaptation is essential.



The left heatmap in Figure~\ref{fig:model_comparison1} reveals critical performance variations across the 22 fine-grained diagnostic labels, with columns representing subcategories and rows denoting different models.  Baseline models exhibit catastrophic failures for clinically significant but low-frequency conditions, particularly for \textit{dark-red tongue} (2.15\% prevalence) where all baselines show $\text{F1} < 0.25$ due to insufficient rare-class representation, and for \textit{gray-black coating} (3.35\%). In contrast, MIRNet-Boosting elevates these to 0.68 and 0.71 F1 respectively through targeted rare-label handling.  Crucially, MIRNet maintains balanced performance across all four diagnostic dimensions, achieving average F1 scores of 0.81 for tongue color, 0.77 for tongue shape, 0.76 for coating property, and 0.84 for coating color, compared to baseline averages of 0.59, 0.43, 0.51, and 0.68 respectively. 

The right heatmap in Figure~\ref{fig:model_comparison1} evaluates dimension-level failures by treating the four diagnostic categories as independent label families, where a missed detection occurs when an image contains at least one true sub-label in a dimension but the model predicts none. Baseline models exhibit severe missed detections, particularly in the tongue shape dimension where maximum misses reach 209 cases. MIRNet-Boosting reduces this to 14 misses, representing a 93.3\% reduction. Similarly, tongue color misses decrease from the baseline range of 36-98 cases to just 18 cases in MIRNet-Boosting, while coating property misses approach zero in MIRNet variants compared to 3-33 in baselines. 

\subsection{Ablation Study}


To isolate the impact of MIRNet's core components, we evaluate three ablated variants:

(1) \textbf{MIRNet\textsubscript{-C}} (Constraint Removal): Removing clinical knowledge integration led to a notable degradation: Example‑F1 and Micro‑F1 both declined by 3.2\%, and Macro‑F1 dropped by 4.4\%, highlighting impaired label consistency.

(2) \textbf{MIRNet\textsubscript{-G}} (GAT-to-MLP Replacement): Substituting the graph attention decoder with a simple MLP classifier caused a 3.2\% loss in Macro‑F1, an 8.1\% drop in recall, and the steepest precision decline (3.1\%) among all variants—confirming the indispensability of GAT for capturing label dependencies.

(3) \textbf{MIRNet\textsubscript{-P}} (Pretraining Removal): Skipping MAE pretraining inflicted the most severe performance hit: Macro‑F1 fell by 23.0\% and Macro Recall collapsed by 29.0\%. This stark deterioration underscores the critical role of pretraining in mitigating annotation scarcity.


Overall, the ablations show complementary roles: clinical constraints deliver the largest overall gains by preventing inconsistent labels; pretraining is most crucial for rare classes, boosting Macro Recall; and the GAT decoder preserves the precision/recall balance, with its removal disproportionately hurting recall and lowering Macro-F1.

\section{Conclusion}

In this work, we introduce MIRNet, a unified framework that couples self-supervised pretraining with constrained graph-based reasoning for tongue-image diagnosis. By combining masked autoencoders, a label co-occurrence graph, and clinically motivated constraints, MIRNet addresses annotation scarcity, label imbalance, and prediction plausibility. On the newly curated TongueAtlas-4K dataset, MIRNet achieves state-of-the-art performance, with especially strong gains on rare labels. Ablation studies substantiate the contribution of each component. Although developed for tongue diagnosis, the framework generalizes naturally to broader medical imaging tasks. Future work will incorporate multi-modal signals and evaluate deployment within clinical workflows to  improve robustness, reliability, and interpretability.

\section*{Acknowledgments}
This work received the following support. The work of Shufeng Kong and Zibin Zheng was partially supported by the SYSU–MUCFC Joint Research Center (Project No. 71010027). The work of Caihua Liu was partially supported by the National Natural Science Foundation of China (Category C; Grant No. 62506090) and the Humanities and Social Sciences Youth Foundation of the Ministry of Education of the People’s Republic of China (Grant No. 21YJC870009). The work of Hao Liang was partially supported by the National Key R\&D Program of China (Grant No.2024YFC3505400) and the Science and Technology Innovation Program of Hunan Province (Grant No. 2022RC1021). The work of Yulong Zhang was partially supported by the Central Funding for the Flagship Chinese–Western Medicine Collaboration (Oncology) Subspecialty Construction Project, the Guangdong Province Basic and Applied Basic Research Fund (Grant No. 2023A1515220179), and the Guangdong Provincial Bureau of Traditional Chinese Medicine (TCM) Scientific Research Project (Grant No. 20231070).
\bibliography{aaai2026}

\clearpage  
\pagebreak[4]  

\end{document}